\documentclass[10pt, a4paper]{article}

\usepackage[final]{lrec2026}
\usepackage{url}
\usepackage{array, booktabs, makecell}
\usepackage{tabularx}
\usepackage{threeparttable}
\usepackage{adjustbox}
\usepackage{ragged2e}
\usepackage{xcolor}     
\usepackage{hyperref}  
\usepackage[capitalize]{cleveref}

\hypersetup{
    colorlinks=true,           
    allcolors=blue!50!black, 
}

\usepackage{expex}   
\usepackage{tikz}
\usetikzlibrary{arrows.meta, positioning, shapes}

\usepackage{graphicx} 

\usepackage{subcaption}
\usepackage{caption}
\usepackage{placeins}
\usepackage{dblfloatfix}
\usepackage{kotex}
\usepackage{multirow}

\lingset{
  aboveexskip=0.5ex,
  belowexskip=1ex,
  interpartskip=0ex
}

\title{EVOKE: Emotion Vocabulary Of Korean and English}

\name{Yoonwon Jung$^1$, Hagyeong Shin$^2$, Benjamin K. Bergen$^1$} 

\address{$^1$Department of Cognitive Science, University of California San Diego, USA\\
$^2$Work done at Department of Linguistics, University of California San Diego, USA\\
         \{y5jung, hashin, bkbergen\}@ucsd.edu\\}

\abstract{
This paper introduces EVOKE (Emotion Vocabulary of Korean and English), a Korean-English parallel dataset of emotion words. The dataset offers comprehensive coverage of emotion words in each language, in addition to many-to-many translations between words in the two languages and identification of language-specific emotion words. The dataset contains 1,426 Korean words and 1,397 English words, and we systematically annotate 819 Korean and 924 English adjectives and verbs. We also annotate multiple meanings of each word and their relationships, identifying polysemous emotion words and emotion-related metaphors. The dataset is, to our knowledge, the most systematic and theory-agnostic dataset of emotion words in both Korean and English to date. It can serve as a practical tool for emotion science, psycholinguistics, computational linguistics, and natural language processing, allowing researchers to adopt different views on the resource reflecting their needs and theoretical perspectives. The dataset is publicly available at \url{https://github.com/yoonwonj/EVOKE}.
 \\ \newline  }

\begin{document}

\maketitleabstract

\section{Introduction}

Emotion words reveal how humans conceptualize and communicate emotional experiences through language \cite{wierzbicka1999emotions}. They also provide a foundation for lexicon-based sentiment analyses and emotion detection \cite{liapis2025enhancing,raji2020sparks}. Yet, emotion words are diverse and notoriously difficult to identify \cite{fehr1984concept, wierzbicka1999emotions, wu2025emotion}. While some words transparently denote emotional states (e.g., \textit{sadness}, \textit{anger}), others evoke or reflect bodily sensations, cognitive evaluations, or physiological or behavioral expressions that may or may not indicate emotional states proper (e.g., \textit{tense}, \textit{captivating}, \textit{blushing}). Consequently, existing datasets of emotion words vary in how they define and annotate emotion words, reflecting a range of theoretical positions and methodological frameworks \cite{clore1987psychological,fehr1984concept, hong2009establishing, johnson1989language, lee2006classification, park2005making}.

A maximally useful dataset of emotion words in a language would have a few features. First, it would be comprehensive--covering all words that could be, on any theory, emotion words. Second, it would explicitly identify selection criteria that establish whether words are or are not emotion terms, and would allow users to apply their preferred criteria to create bespoke theory-driven views on the data. However, existing datasets generally cover subsets of the possible emotion words in any language. Moreover, they often apply different criteria to identify emotion words based on specific theoretical frameworks, which makes it challenging to integrate or compare datasets \cite{fehr1984concept, baron2010emotion, clore1987psychological, hong2009establishing, johnson1989language, lee2006classification, park2005making}. 

Cross-linguistic datasets extend the value of monolingual resources, as comparisons of emotion words can reveal emotions that may be shared or specific across cultures \cite{jackson2019emotion, russell1991culture}. To date, however, the coverage of words in cross-linguistic datasets is limited. Because meanings do not always translate one-to-one across languages \citep{catford1965linguistic, thompson2020cultural}, a word in one language may not have any direct translation in other languages (and thus create lexical gaps), or might align with multiple words in another language. For example, while English \textit{dismay} does not have an exact translation in Korean, English \textit{sad} can correspond to multiple, subtly distinct Korean words (i.e., \textit{서럽다} (\textit{sŏrŏpta})\footnote{To be sad and depressed \citep{krdictseo}.} and \textit{슬프다} (\textit{sŭlpŭta})\footnote{Sad and sorrowful enough to make one cry \citep{krdictseul}.}, each of which also has several different English correspondents of its own. This underscores the need for cross-linguistic datasets of emotion words with comprehensive lexical coverage in each of the languages and many-to-many translational mappings.

Lastly, emotion words are complex in that they are often polysemous. For example, Korean \textit{부끄럽다} (\textit{pukkŭrŏpta}) encompasses the meaning of both \textit{shy} or \textit{shameful}. Other polysemous senses are often metaphorical. One example is the emotional and spatial senses of \textit{low} in English. Metaphorical relations among senses of emotion words, and polysemy more generally, reveal how emotion semantics are organized within and across languages \citep{kovecses2003metaphor, sauciuc2009role}, yet they are rarely documented in emotion word datasets.

In light of these gaps, we present a dataset of emotion words in two languages with (1) comprehensive lexical coverage and annotations by native speakers based on several different theories to accommodate diverse views and uses, (2) many-to-many translational mappings established by bilingual experts using existing lexicographic resources, and (3) identification of polysemous words and metaphorical relations among word senses.

\section{Background}
\subsection{Theories Defining Emotion Words}
\label{sec:prior_emotion_word}

Psychological and linguistic work has produced varied methods to define emotion words. Some early studies relied on unguided intuition, with raters judging words as emotional or not without explicit criteria \citep{fehr1984concept, storm1987taxonomic}. More recent work uses acceptability judgments on emotion words, placing them in frame sentences for expressing emotions like ``I feel X'' or ``I am X'' to evaluate whether they sound acceptable in those contexts \citep{baron2010emotion, clore1987psychological, johnson1989language}. Another recent approach uses exclusion criteria, asking if a word describes bodily sensation or epistemic state without emotionality, to filter out some obvious non-emotion words \citep{baron2010emotion, johnson1989language, park2005making}. 

Research on Korean emotion words classifies emotion-related predicates into different types: state-oriented (a state experienced by an animate subject), evaluation-oriented (properties of external entities), and expression-oriented (behavioral expressions and responses) \citep{hong2009establishing,lee2006classification,park2005making}. The distinction between stative and evaluative predicates is closely tied to causality, echoing the proposal to distinguish externally caused emotions (e.g., ``I am saddened by his death'') from those concerning internal goals (e.g., \textit{desire}) \citep{johnson1989language}.

\subsection{Cross-linguistic Studies}
The majority of existing cross-linguistic studies of emotion words prioritize the prototypicality of emotion over exhaustiveness, focusing on small sets of emotion terms presumed to be universal (typically fewer than 30) across large numbers of languages \citep[e.g.,][]{list2018clics2}, or utilize restricted sets of words from bilingual studies \citep[e.g.,][]{bromberek2021cross, tang2023embodiment}. These approaches limit the exploration of finer-grained distinctions of emotion words within a language. 

Another line of research highlights culture-specific emotion words, captured as words lacking equivalents in other languages (i.e., \textit{lexical gaps}). For example, Korean words \textit{정} (\textit{chŏng}) or \textit{한} (\textit{han}) lack equivalents in English, and are expressible only with longer descriptions, like ``long-lasting affection and caring based on perceived closeness'' \citep{choi2002effects} or ``Unresolved resentment, grief, and anger, or a negative and long-lasting emotion encapsulating the grief of historical memory'' \citep{kim2017korean}, respectively. Cross-linguistic lexical gaps signal how different language communities represent and express emotions differently \cite{lomas2018experiential,Lupyan2012chapter,rissman2023gaps,winawer2007russian}. However, they are often discovered in small numbers through case studies (e.g., \citealp[]{choi2002effects, kim2017korean, schmidt1999korean}), and are rarely identified at scale within a broader lexicon of emotion in one language in relation to other languages.

\subsection{Semantic Relations and Metaphors}
The meanings of polysemous emotion words are often connected through metaphorical extension from one meaning to other meanings \cite{bartsch2002generating}. Comparing patterns of such emotion-related metaphors across languages can reveal how emotional meanings are structured similarly or differently across cultures \cite{kovecses2003metaphor}. Identifying these multiple senses thus provides a basis for cross-linguistic studies on emotion semantics.

Existing annotation work on polysemous words mainly targets resolving general lexical ambiguities, not emotion words specifically \citep{haber2024polysemy, passonneau2010word, rumshisky2008polysemy}. Similarly, cross-linguistic studies of emotion-related metaphors often focus on a small set of recurrent metaphorical mappings (e.g., ``emotion as force'' or ``anger as a container''; \citealp{lakoff1987cognitive, sauciuc2009role, turker2013corpus}), rather than examining metaphorical structures across a wider variety of emotion words. These gaps leave open questions about how emotion words develop multiple meanings and metaphorical extensions, and how such patterns distribute within and across languages.

\begin{figure*}[t]
\centering
\resizebox{\textwidth}{!}{
\begin{tikzpicture}[
  font=\sffamily,
  korword/.style={rectangle,draw=black,rounded corners,fill=white,
                  minimum width=4.6cm,minimum height=0.8cm,align=center},
  engword/.style={rectangle,draw=black,rounded corners,fill=white,
                  minimum width=2.1cm,minimum height=0.8cm,align=center},
  block/.style={rectangle,rounded corners,fill=gray!35,
                minimum width=10.5cm,minimum height=5cm,align=center,inner sep=6pt},
  side/.style={rectangle,rounded corners,fill=gray!35,
               minimum width=3.5cm,minimum height=1.8cm,align=center,font=\bfseries},
  arrow/.style={-{Latex[length=3mm,width=2mm]},thick},
  conn/.style={thick}
]

\node[block] (db) at (0,0) {};

\node[font=\bfseries] at (0,2.0) {Korean–English Mappings};
\node[font=\normalsize] at (0,-2.0) {Many-to-many translational mappings};

\node[font=\bfseries] at (-4.5,0) {Kor};
\node[font=\bfseries] at ( 4.5,0) {Eng};

\node[korword,anchor=west] (kor1) at (-4.0, 1.0) {\textit{수치스럽다} (\textit{suchi'sŭrŏpta})};
\node[korword,anchor=west] (kor2) at (-4.0, 0.0) {\textit{부끄럽다} (\textit{pukkŭrŏpta})};
\node[korword,anchor=west] (kor3) at (-4.0, -1.0) {\textit{수줍다} (\textit{suchupta})};

\node[engword,anchor=east] (eng0) at (4.0, 0.0) {\textit{bashful}};
\node[engword,anchor=east] (eng1) at (4.0, 1.0) {\textit{shameful}};
\node[engword,anchor=east] (eng2) at (4.0, -1.0) {\textit{shy}};

\draw[conn] (kor2.east) -- (eng0.west);
\draw[conn] (kor1.east) -- (eng1.west);
\draw[conn] (kor2.east) -- (eng1.west);
\draw[conn] (kor3.east) -- (eng2.west);
\draw[conn] (kor2.east) -- (eng2.west);

\node[side] (leftA)  at (-8.2,0) {Annotations for\\Korean Words};
\node[side] (rightA) at ( 8.2,0) {Annotations for\\English Words};

\draw[arrow] (db.west) -- (leftA.east);
\draw[arrow] (db.east) -- (rightA.west);

\end{tikzpicture}
}
\caption{The structure of the Korean--English parallel emotion word dataset. Words in both languages are connected through many-to-many translational mappings, with annotations for Korean and English words.}
\label{fig1}
\end{figure*}

\section{Objective of the Dataset}
We introduce EVOKE (Emotion Vocabulary Of Korean and English), a Korean-English parallel dataset of emotion words with comprehensive coverage in both languages plus cross-linguistic mappings (publicly available at \url{https://github.com/yoonwonj/EVOKE}). The dataset is constructed by compiling lists of words to annotate from prior studies on and datasets of emotion words. Those words are then annotated according to multiple criteria used in earlier studies to investigate their characteristics as emotion words. The dataset also provides many-to-many translations of words between Korean and English, which enables discovering emotion concepts specific to each language. Lastly, the dataset identifies polysemous words and annotates the relationships between their senses, enabling the identification of emotion-related metaphors. 

We focus on Korean and English for several reasons. English serves as a well-studied reference point, yet there does not yet exist a comprehensive resource with structured annotations like the one constructed here. Korean, meanwhile, is spoken in distinct cultural settings, and contains purportedly unique, culture-specific emotion terms (e.g., \textit{정} (\textit{chŏng}), \textit{한} (\textit{han})) that lack direct English equivalents \citep{choi2002effects, kim2017korean}. Comparison of the two allows for the investigation of both shared and culture-specific aspects of emotion lexica. Additionally, the research team's expertise in linguistics, cognitive science, and knowledge as native speakers of English and Korean makes it feasible to construct a reliable dataset.

\begin{table*}[t]
\centering
\renewcommand{\arraystretch}{1.15}
\begin{adjustbox}{width=\textwidth}
\begin{tabular}{p{1.2cm} p{4.5cm} p{11.1cm}}
\toprule
\textbf{ID} & \textbf{Label} & \textbf{Annotation Question} \\
\midrule
\multicolumn{3}{l}{\textbf{Part 1. Acceptability judgments}} \\
\midrule
\textsc{acpt}1 & 1st person ``feel''  & Does ``I feel X'' sound acceptable?\\
\textsc{acpt}2 & 3rd person plural ``feel'' & Does ``They feel X'' sound acceptable?\\
\textsc{acpt}3 & Inanimate subject ``feel'' & Does ``It feels X'' sound acceptable (inanimate ``it'')?\\
\textsc{acpt}4 & 1st person ``am''  & Does ``I am X'' sound acceptable?\\
\midrule
\multicolumn{3}{l}{\textbf{Part 2. Semantic experiencer judgments}} \\
\midrule
\textsc{exp}5 & Subjectivity of experience & In ``It feels X'', can the sentence express what ``it'' experiences?\\
\textsc{exp}6 & Evaluation as experience & In ``It feels X'', can the sentence express my evaluation of ``it''?\\
\textsc{exp}7 & Caused experience & In ``I am X'', does X denote a caused state of its associated noun/verb?\\
\textsc{exp}8 & Causing experience & In ``I am X'', does X denote a causing state of its associated noun/verb?\\
\midrule
\multicolumn{3}{l}{\textbf{Part 3. Exclusion criteria}} \\
\midrule
\textsc{excl}9 & Pure bodily sensation & Does X describe a pure bodily/physical sensation?\\
\textsc{excl}10 & Behavioral expression & Does X denote a behavioral expression?\\
\textsc{excl}11 & Pure epistemic state & Does X denote a non-emotional epistemic state?\\
\midrule
\multicolumn{3}{l}{\textbf{Part 4. Multiple meanings}} \\
\midrule
\textsc{poly}12 & Additional Meaning & Does X have another distinct meaning to annotate separately? \newline $\rightarrow$ If yes: create a new row for that sense and complete annotations in Parts 1–3.\\
\textsc{poly}13 & Distinctiveness & Are the two senses of X in different domains?\\
\textsc{poly}14 & Relatedness & Are the two senses of X systematically related?\\
\bottomrule
\end{tabular}
\end{adjustbox}
\caption{Codebook for the annotation scheme. Part 1 collects acceptability judgments of the target word in four different sentences. Part 2 includes two follow-ups for \textsc{acpt}3 (\textsc{exp}5--6) and \textsc{acpt}4 (\textsc{exp}7--8) each. Part 3 applies exclusion criteria to filter relatively obvious non-emotion terms. Part 4 documents polysemy and links senses via distinctiveness and relatedness judgments. All questions used two binary labels (\texttt{acceptable}/\texttt{unacceptable} in Part 1, \texttt{yes}/\texttt{no} in Parts 2--4), with an additional option to choose \texttt{unsure}. \textsc{acpt}1, \textsc{acpt}2, \textsc{acpt}3, \textsc{exp}5, and \textsc{exp}6 were only applied to adjectives, and the rest of the others were applied to both verbs and adjectives, with modifications for verb annotations. Annotation criteria were translated into Korean for Korean word annotations with appropriate modifications applied.}
\label{tab_annotation}
\end{table*}

\section{Dataset Construction}
The dataset consists of three separate components (see Figure~\ref{fig1}): (1) Korean-English mappings, (2) annotations for Korean words, and (3) annotations for English words.

\subsection{Word Selection and Translation}
Candidate words were gathered from previous work on emotion words in English \cite{baron2010emotion, morgan1988structure, storm1987taxonomic} and Korean \cite{jeon2022user, park2005making, rhee2013measuring}.

To construct translational mappings (illustrated in Figure~\ref{fig1}), two Korean-English bilingual speakers judged whether each word had translational equivalents in the other language. A manual translation approach was adopted to capture nuanced and precise meanings that are difficult to obtain through automatic translation. The translators consulted two bilingual Korean-English dictionaries (one Korean-to-English and one English-to-Korean) for translation \cite{krdict, naverdict}. In addition, definitions from Korean \cite{skld} and English monolingual dictionaries \cite{oed, cambdict} were compared to ensure accurate and semantically rich matching, especially when translations provided in bilingual dictionaries varied across sources or when forward and backward translation yielded inconsistent results.

A word is considered to have a translational equivalent when it can be translated to a single word form (i.e., one or more words exist in both languages to denote the same meaning). Accordingly, lexical gaps were identified when (1) bilingual dictionary queries for words in one language only yielded multi-word expressions or idiomatic phrases in the other language,\footnote{Hyphenated forms were treated as single lexical items (e.g., \textit{cliché-ridden}). Korean words that may appear either as single words or with internal spaces were treated as single-word entries and were not considered as lexical gaps (e.g., \textit{사려 깊은} (\textit{saryŏkipŭn}) `considerate').} or when (2) back-translation produced inconsistent results. In the latter cases, the translators consulted additional native speakers and monolingual dictionaries to determine whether the dictionary equivalent reflected a true translational match that fully captured the meaning of the query word.

Moreover, lexical gap was defined as the absence of a word for a concept in one language when such a word exists in the other (e.g., \citealp{janssen2004multilingual, li2024translation}). Therefore, gaps arising at the morphosyntactic level were not included, as those do not indicate the absence of conceptual or semantic content \cite{ivir1977lexical, lomas2018experiential, wierzbicka1999emotions}.\footnote{Some Korean verbs and nouns are systematically translated into English using \textit{be-} constructions (verb) and \textit{being-} constructions (noun). This creates gaps in verbs and nouns for English relative to Korean in a systematic way. See \citealt{bentivogli2000coping} for additional examples and discussion of morphosyntactic gaps.}

\subsection{Annotation Objective and Process}
We annotated words whose part-of-speech is identified as adjectives and verbs in each language. Adjectives linked to identifiable lexical roots\footnote{A form commonly perceived as a base form without its morphological derivation.}, either nouns or verbs, were annotated as noun--adjective or verb--adjective pairs. Adjectives without such identifiable lexical roots, along with all verbs, were annotated independently. Nouns were not annotated because the sentence contexts most commonly used for acceptability judgments in the literature are designed around predicative uses of words \cite{baron2010emotion, clore1987psychological, johnson1989language}, and several existing annotation criteria target semantic properties specific to predicates (see Section~\ref{sec:exp}; \citealp{hong2009establishing, lee2006classification, park2005making}).

We recruited three native English speakers and three native Korean speakers as annotators\footnote{Annotators received course credits as compensation.}, who had backgrounds in cognitive science or linguistics. Annotators completed one week of training followed by ten weeks of annotation. During weekly meetings, researchers and annotators discussed questions and edge cases. Those discussions resulted in modifications to the annotation guidelines meant to improve the robustness of subsequent annotations. 

Annotators made binary judgments (\texttt{acceptable}/\texttt{unacceptable}; \texttt{yes}/\texttt{no}) for each annotation criterion, with the option to choose \texttt{unsure}. We chose binary judgments over graded ratings because some of the questions are intrinsically categorical. Binary ratings also avoid subjectivity introduced by continuous scales (e.g., differential use of Likert scales per annotator) and allow for easier combination of annotation results from the different criteria to select words that qualify as emotion words (see Section~\ref{sec:selection} for an example of a use case).

Items marked \texttt{unsure} were reviewed in weekly meetings. Cases that could be resolved through group discussion were annotated again with binary labels. However, given the subjective nature of emotion word judgments, we did not force consensus. When annotators continued to find it difficult to make a binary decision, we kept the \texttt{unsure} label.

The dataset was split so that each annotator received a unique 30\% of the words, while an additional 10\% of the words were assigned to all three annotators for inter-rater reliability assessment. 

\subsection{Annotation Criteria}
\label{sec:annot_criteria}
The annotation criteria used features of emotion words identified from the existing literature. Those criteria were selected to include or exclude candidate words from the category of emotion words. The full list is presented in Table~\ref{tab_annotation}.

Adjectives were annotated on all 14 criteria, while verbs were annotated on 9, excluding \textsc{acpt}1, \textsc{acpt}2, \textsc{acpt}3, \textsc{exp}5, and \textsc{exp}6. In addition, \textsc{acpt}4 was modified to suit part-of-speech differences. Because the criteria differ by language and by part-of-speech, Table~\ref{tab_annotation} reports the annotation scheme for English adjectives, which encompasses all 14 of the criteria we devised. For Korean annotations, the criteria were translated and adapted to Korean syntax and usage patterns. Details of the modifications to Korean word annotations are in Appendix~\ref{sec:appendix_kor_annot}.

\subsubsection{Part 1: Acceptability Judgments}

The first part of the annotation collected acceptability judgments to elicit native speakers' usage patterns of the words.\footnote{We ground the acceptability judgments in pragmatic acceptability instead of grammatical acceptability.} Annotators judged whether target words sound \texttt{acceptable} or \texttt{unacceptable} in given sentences, with an option to choose \texttt{unsure}. The sentences served as standardized contexts for evaluating each emotion word's appropriateness for expressing emotion. When the word alone sounded unnatural, annotators could suggest an alternate phrasal structure, which was then adopted for the annotation (e.g., ``I feel cared'' could be replaced by ``I feel cared for'').\footnote{This was in response to differences between English, which commonly uses phrasal verbs, and Korean, which instead tends to use morphological derivation.} 

The sentences for acceptability judgments were selected based on their previous adoption as inclusion criteria for emotion words \cite{baron2010emotion, clore1987psychological, johnson1989language}, from psychological and linguistic studies of emotion \cite{hong2009establishing, liu2016emotion, lee2006classification, niedenthal2008emotion,paul2020towards}.

\textbf{\textsc{acpt}1: 1st person ``feel'', \textsc{acpt}2: 3rd person plural ``feel'', \textsc{acpt}3: Inanimate subject ``feel''}
Three sentence contexts based on ``I feel X'' but with varying subjects were used to assess whether words activate subjective feelings of emotion \cite{lee2006classification,niedenthal2008emotion,paul2020towards}: ``I feel X'' (\textsc{acpt}1: 1st person ``feel''), ``They feel X'' (\textsc{acpt}2: 3rd person plural ``feel''), and ``It feels X'' (\textsc{acpt}3: Inanimate subject ``feel''), where X is a target word. Participants were instructed to interpret ``I'' and ``they'' as animate experiencers (1st person and 3rd person plural), and ``it'' as an inanimate object. If a word denotes a subjective feeling, it is expected to be judged as acceptable with animate subjects, but not with inanimate ones \cite{baron2010emotion, clore1987psychological, dowty1991thematic, lee2006classification, liu2016emotion}.

\textbf{\textsc{acpt}4: 1st person ``am''}
``I am X'' (\textsc{acpt}4: 1st person ``am'') was used to contrast with ``I feel X'' (1st person ``feel'') and capture words denoting states versus evaluations. For example, \textit{caring} fits ``I am caring'' but not ``I feel caring''. This contrast helps reveal whether words refer to internal states or external evaluations. However, some words (e.g., \textit{pretty}) can still be acceptable in both ``I feel X'' and ``I am X''. Follow-up questions in Section~\ref{sec:exp} further clarify these cases \cite{clore1987psychological, hong2009establishing, lee2006classification, park2005making}.

\subsubsection{Part 2: Semantic Experiencer Judgments}
\label{sec:exp}

The second part introduced semantic experiencer judgments as follow-up questions to \textsc{acpt}3 and \textsc{acpt}4. These questions aimed to handle patterns in acceptability judgments like those for \textit{pretty}. Specifically, these questions helped determine whether words describe subjective experiences or external evaluations, reflecting the contrast between stative and evaluative words \cite{choi2008type, hong2009establishing, lee2006classification}.

\textbf{\textsc{exp}5: Subjective experience, \textsc{exp}6: Evaluation}
Annotators judged whether the word in ``It feels X'' (\textsc{acpt}3: Inanimate subject ``feel'') could express what the inanimate subject ``it'' experiences (\textsc{exp}5), and whether it could express the speaker’s evaluation of ``it'' (\textsc{exp}6).  These follow-up judgments clarify whether \textsc{acpt}3 acceptability stems from evaluative readings. For instance, ``it feels merciless'' may sound acceptable if ``it'' refers to a fighter jet, but such usage reflects evaluation rather than the inanimate subject’s feelings. These follow-up questions ensure the frame captured the intended semantics and help sort out such exception cases.

\textbf{\textsc{exp}7: Caused, \textsc{exp}8: Causing}
Follow-up questions for \textsc{acpt}4 (1st person ``am'') were designed to complement the distinction between experiential and evaluative words, following the ``caused emotion'' distinction of \citet{johnson1989language}. Annotators were asked to judge whether the target word describes a ``caused'' (\textsc{exp}7) or ``causing'' (\textsc{exp}8) state within the ``I am X'' sentence context. Korean annotators were allowed to adjust the subject to consider additional naturalistic contexts, incorporating the annotators' suggestions for more comprehensive evaluation.

To guide these judgments, adjectives were paired with their identifiable root nouns or verbs (e.g., \textit{sadness}–\textit{sad}), and were asked to judge whether each adjective is a caused state or a cause of the paired root nouns or verbs. For example, annotators judged whether \textit{sad} refers to the caused state of \textit{sadness} or its causing state. For the remaining adjectives, annotators were asked to imagine hypothetical noun forms. 

These causality judgments clarify whether a word denotes an internal state or an external evaluation: internal states typically correspond to caused forms (e.g., \textit{depressed}), while evaluations often align with causing forms (e.g., \textit{depressing}). A word may encompass both caused and causing meanings, or words may represent each meaning separately. Moreover, words with similar meanings in different languages may capture these caused and causing meanings differently. Such differences could help reveal varying semantic properties of emotion words across languages.

\subsubsection{Part 3: Exclusion Criteria}
The third part applied exclusion criteria \cite{park2005making, baron2010emotion, lee2006classification}. While emotions involve coordinated mind–body processes, most theories distinguish them from pure bodily sensations (e.g., \textit{thirsty}) or cognition (e.g., \textit{enlightened}) \citep{anderson2014framework,johnson1989language,lee2006classification,paul2020towards}. Many also separate emotional states from physical expressions (e.g., \textit{cry}), which are byproducts rather than core states \citep{anderson2014framework,johnson1989language,hong2009establishing,lee2006classification,paul2020towards}. These judgments are used to guide the exclusion of non-emotion words.

Annotators judged whether each target word describes pure bodily or physical sensations (\textsc{excl}9), behavioral expressions (\textsc{excl}10), or pure epistemic states (\textsc{excl}11). These criteria were designed to flag words that primarily fell into these categories for potential exclusion from the set of emotion words. This procedure provides a direct way to filter out non-emotion terms according to widely shared theoretical assumptions, still allowing researchers to retain those as emotion words if desired.

\subsubsection{Part 4: Multiple Meanings}

\textbf{\textsc{poly}12: Additional meaning}
The fourth part of the annotation was introduced to capture whether each word had other meanings distinct from its primary emotional meaning (\textsc{poly}12). When annotators identified an additional meaning, they were instructed to complete annotations in Parts 1--3 for the additional meaning identified. To further specify semantic relationships among multiple senses, two follow-up questions assessed the relationship between the identified meanings.

\textbf{\textsc{poly}13: Distinctiveness, \textsc{poly}14: Relatedness}
Annotators judged whether the two senses concern different domains or concepts (Distinctiveness; \textsc{poly}13), and whether they are connected in a systematic way (Relatedness; \textsc{poly}14). For example, the word \textit{low} exemplifies a prototypical metaphorical relationship. Its physical and emotional meanings belong to distinct domains yet remain systematically linked through the conceptual mapping between spatial position and emotional valence. In contrast, \textit{blue} demonstrates a looser connection in that its color and emotional senses are distinct, but their association lacks systematic grounding. Although the metaphorical status of \textit{blue} is debatable, it does not exhibit the systematicity typical of metaphors.

\pex
\a ``The ceiling is low.'' $\rightarrow$ \textsc{height}
\a ``I am feeling low today.'' $\rightarrow$ \textsc{feeling}
\a $\rightarrow$ \texttt{POLY}13: \texttt{yes}, \texttt{POLY}14: \texttt{yes}
\xe

\pex
\a ``The sky is blue.'' $\rightarrow$ \textsc{color}
\a ``I feel blue.'' $\rightarrow$ \textsc{feeling}
\a $\rightarrow$ \textsc{poly}13: \texttt{yes}, \textsc{poly}14: \texttt{no}
\xe

These dimensions derive from conceptual metaphor theory, and specifically the Metaphor Identification Protocol \cite{declercq2023coding,group2007mip, kovecses2010metaphor}, to annotate words that could have emotional meaning through metaphorical extension (e.g., \textit{low}). The annotation results could help reveal how emotional meanings emerge and relate to other meanings.

\section{Dataset Analysis and Evaluation}
\subsection{Word characteristics}

\subsubsection{Part-of-Speech Statistics}
Summary word statistics are in Table~\ref{tab4}. A total of 1,426 Korean words and 1,397 English words were included. Translation equivalents were identified for all of these, forming many-to-many mappings across Korean and English (Figure~\ref{fig1}).

\begin{table*}[ht!]
\centering
\renewcommand{\arraystretch}{1.2}
\begin{tabular}{lcccc}
\toprule
 & \textbf{Nouns} & \textbf{Adjectives} & \textbf{Verbs} & \textbf{Total} \\
\midrule
\textbf{Korean words} & 591 & 606 & 229 & 1,426 \\
\quad With translation(s) in English 
& 574 (97.12\%) & 587 (96.86\%) & 213 (93.01\%) & 1,374 (96.35\%) \\
\quad Without translation(s) in English 
& 17 (2.88\%) & 19 (3.14\%) & 16 (6.99\%) & 52 (3.65\%) \\
\midrule
\textbf{English words} & 508 & 671 & 218 & 1,397 \\
\quad With translation(s) in Korean 
& 495 (97.44\%) & 630 (93.89\%) & 213 (97.71\%) & 1,338 (95.80\%) \\
\quad Without translation(s) in Korean 
& 13 (2.56\%) & 41 (6.11\%) & 5 (2.29\%) & 59 (4.22\%) \\
\bottomrule
\end{tabular}
\caption{The statistics of Korean and English lexical entries with and without translational equivalents. Words without translational equivalents indicate lexical gaps. Polysemous words with either (1) multiple parts of speech, or (2) one of the meanings having translations in the other language but not in the other, were counted as separate lexical entries. See Appendix~\ref{sec:appendix_poly} for policies regarding counting polysemous word entries.}
\label{tab4}
\end{table*}

\subsubsection{Translational Equivalence}
The majority of words had corresponding translations in the other language. For Korean words, 574 nouns (97.12\%), 587 adjectives (96.86\%), and 213 verbs (93.01\%) had one or more translational equivalents in English. For English words, 495 nouns (97.44\%), 630 adjectives (93.89\%), and 213 verbs (97.71\%) had one or more corresponding words in Korean.

The analysis of translational equivalents revealed a relatively similar degree of mapping complexity between the two languages. Korean words had an average of 1.61 translational equivalents in English (SD = 0.93), and English words had a nearly identical average of 1.65 translational equivalents in Korean (SD = 1.15). A total of 472 words in English (33.79\%) and 542 words in Korean (38.01\%) had two or more translational equivalents in the other languages. Some words demonstrated particularly rich translational mappings. The Korean adjective \textit{우울하다} (\textit{uulhata}) exemplified this complexity with six English equivalents: \textit{depressed}, \textit{dismal}, \textit{gloomy}, \textit{low}, \textit{down}, and \textit{blue}, capturing a range of physical and emotional states that Koreans express through a single word.

\subsubsection{Lexical Gaps}
Among Korean words, 17 nouns (2.88\%), 19 adjectives (3.14\%), and 16 verbs (6.99\%) lacked translational equivalents in English. In English, 13 nouns (2.56\%), 41 adjectives (6.11\%), and 5 verbs (2.29\%) had no corresponding translations in Korean. 

These lexical gaps may indicate conceptual gaps, though this is not always the case. For example, some of those words indicate areas where each language expresses emotion concepts uniquely, like \textit{답답하다} (\textit{taptaphata})\footnote{Feeling dissatisfied and uncomfortable when someone's behavior or the situation makes it hard to meet one's expectation \cite{krdictdap,schmidt1999korean}.} \cite{schmidt1999korean}. No single English word is equivalent, nor is there a conventionalized phrasal way to express the concept it denotes. However, other lexical gaps do have phrasal equivalents, like \textit{moody}, which has a multi-word equivalent to ``having ups and downs'' (\textit{기분 변화가 심한} (\textit{kipun byŏnhwaka simhan})) in Korean. Systematically distinguishing these instances from each other is beyond the scope of this paper. However, because the identified lexical gaps include both cases, this dataset can be utilized to further explore empirical questions regarding the relationship between lexical gaps and conceptual gaps and the effects--if any--of the conventionality of multi-word descriptions for emotion concepts.

\begin{figure*}[!ht]
    \centering
    \includegraphics[width=0.95\textwidth]{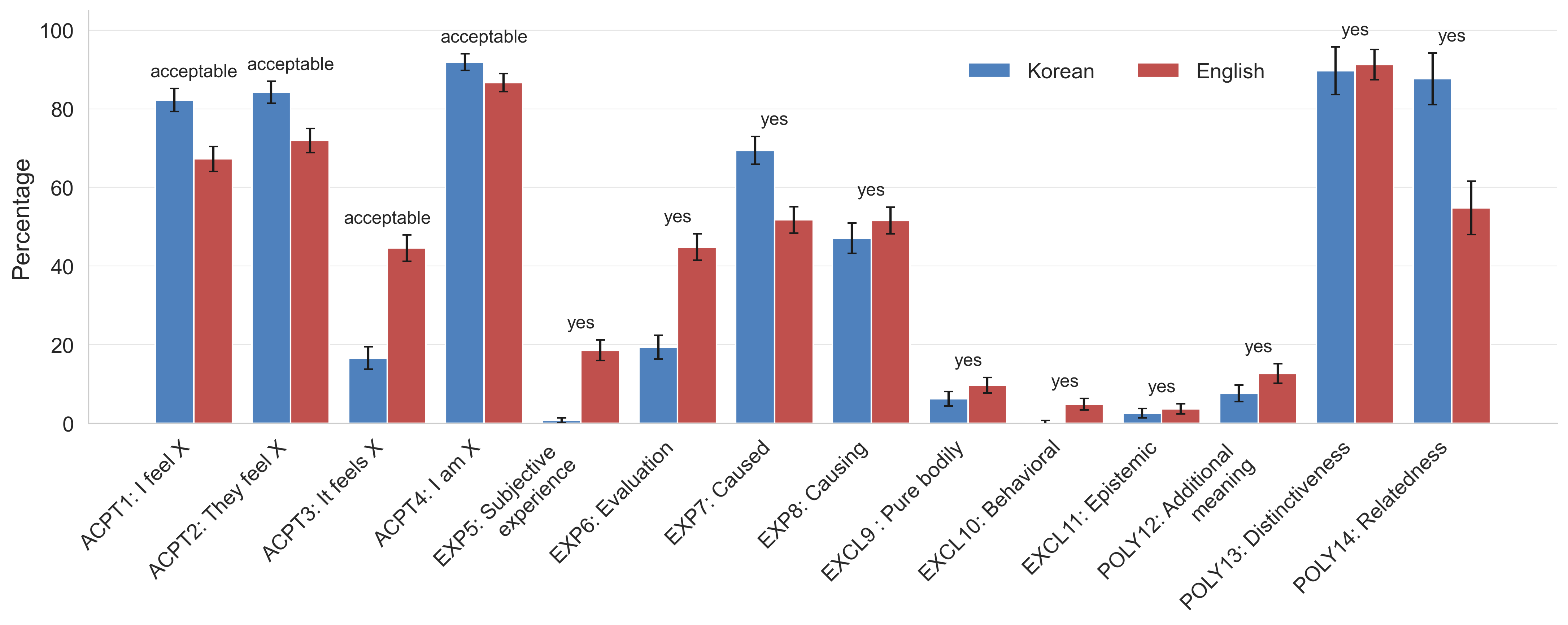}
    \caption{The percentages of annotation values for each annotation criterion for adjectives in Korean and English. The percentages for \textsc{poly}12 indicate the ratio of words annotated as having more than one meaning in each language. The percentages for \textsc{poly}13 and \textsc{poly}14 were calculated among items annotated as having more than one meaning in \textsc{poly}12. Error bars indicate 95\% confidence intervals.}
    \label{fig_annotation_trend}
\end{figure*}

\subsection{Annotation Results}
A total of 819 Korean words (602 adjectives, 217 verbs) and 923 English words (678 adjectives, 245 verbs) were annotated.\footnote{Discrepancies between the number of words annotated and those included in the translational mappings arose from post-hoc corrections to part-of-speech coding and translation mappings.} 

Annotation criteria differed by part of speech, which made averaging results across adjectives and verbs infeasible. We thus report results for adjectives, which constitute the largest portion of the dataset, and use all criteria (see Section~\ref{sec:annot_criteria} for details). Results of verb annotations are provided in Appendix~\ref{sec:annot_verb_result}, and \cref{fig_pairwise_kappa_eng_verb,fig_pairwise_kappa_kor_verb,fig_annotation_trend_verb}.

\subsubsection{Agreement Scores}
Agreement scores were calculated for the annotation criteria in Parts 1--3, as not all words had multiple meanings to be annotated. See Figures~\ref{fig_pairwise_kappa_eng_adj} and \ref{fig_pairwise_kappa_kor_adj} in Appendix for the full pairwise annotation scores.

Overall, Korean annotations showed higher reliability (mean Cohen’s $\kappa = 0.74$) than English annotations (mean $\kappa = 0.60$) in average. This pattern may reflect greater consistency in the features of Korean emotion words or, alternatively, additional training adjustments introduced for the Korean word annotation after the annotation of English words.  

Agreement also varied by criterion and cross-linguistically. Exclusion criteria produced the highest agreement in both languages (mean \textsc{excl}9--11: Korean $\kappa = 0.89$, English $\kappa = 0.82$). Causality criteria showed the lowest agreement across both languages (\textsc{exp}7--8: mean Korean $\kappa = 0.52$, mean English $\kappa = 0.45$), with the ``causing'' criterion (\textsc{exp}8) being particularly challenging (Korean $\kappa = 0.46$, English $\kappa = 0.45$). For the acceptability judgments (\textsc{acpt}1--4), Korean annotators reached substantial agreement (mean $\kappa = 0.73$), while English annotators reached moderate to substantial levels (mean $\kappa = 0.58$). The sub-criteria for subjective experience and evaluation (\textsc{exp}5--6) achieved higher agreement in Korean (mean $\kappa = 0.76$) than in English (mean $\kappa = 0.47$). 

These inter-annotator agreement patterns highlight both the subjective nature of emotion concepts and the theoretical disagreements that surround them. In particular, the relatively low agreement on causality judgments likely reflects the abstract and complex reasoning required. Annotators likely relied on their subjective judgments that relied on personal experiences and their language use. As such, criteria like \textsc{exp}7 (``Caused'') and \textsc{exp}8 (``Causing'') should be applied cautiously in making inclusion or exclusion decisions when defining emotion words.

\subsubsection{Annotation Trend}

A summary of annotation trends for all annotated adjectives is presented in Figure~\ref{fig_annotation_trend}. Annotations in the agreement set were adjudicated using majority vote to produce a single combined dataset for analysis.

Words in both languages exhibited higher acceptance in ``I am'' (\textsc{acpt}4; Korean: 91.83\%, English: 86.85\%) than in ``I feel'' (\textsc{acpt}1; Korean: 82.13\%, English: 67.23\%) or in ``They feel'' (\textsc{acpt}2; Korean: 84.28\%, English: 71.91\%) sentences. Most words were annotated as unacceptable in ``It feels'' (\textsc{acpt}3) in Korean (83.36\%), but not in English (55.14\%). 

Responses for causality criteria were more evenly distributed in both languages (\textsc{exp}7 Korean: 69.18\% \texttt{yes}, English: 51.56\% \texttt{yes}; \textsc{exp}8 Korean: 46.69\% \texttt{yes}, English: 51.82\% \texttt{yes}). Exclusion criteria (\textsc{excl}9--11) showed a highly consistent rejection rate (around 90\%) in both languages. 

More English adjectives had multiple meanings annotated (12.68\%) than in Korean (7.48\%). Cross-linguistic differences also appeared in relatedness judgments (\textsc{poly}14; Korean 87.37\%, English 55.00\%), whereas distinctiveness judgments converged (\textsc{poly}13; Korean 89.47\%, English 92.22\%). Annotation of additional meaning was correlated positively with bodily meanings in both languages, though more strongly in Korean ($r = 0.46$) than English ($r = 0.26$), suggesting a pattern of metaphorical extension from physical to emotional domains. 

\section{Future Applications}

\subsection{Theory-Driven Selection of Words}
\label{sec:selection}
This dataset is theory-agnostic, as it does not commit to a single theoretical stance on how emotion words should be defined. Researchers can apply their preferred annotation criteria to select emotion words that align with their theoretical perspectives and research needs, enabling the curation of experimental stimuli for both emotion research and computational studies of emotion word semantics.

A conservative approach to selecting words from the dataset applies decision criteria from prior studies. Specifically, (1) a word should be acceptable in ``I feel'', ``They feel'', and ``I am'' sentences, but not in ``It feels'' context (Part 1; \texttt{yes} for \textsc{acpt}1, \textsc{acpt}2, \textsc{acpt}4; \texttt{no} for \textsc{acpt}3); (2) if acceptable in ``It feels'', it should express human evaluation rather than an inanimate subject’s experience (Part 2; \texttt{no} for \textsc{exp}5 and \texttt{yes} for \textsc{exp}6); and (3) it should not denote pure bodily sensations, behavioral expressions, or non-emotional epistemic states (Part 3; \texttt{no} for \textsc{excl}9--11).

Applying these criteria yields 425 single-word emotion adjectives in Korean and 317 in English. Examples of excluded words include \textit{mean} (\textit{못되다} (\textit{mottwaeta}) in Korean), which feels unnatural in ``I feel X'', and \textit{clueless} (\textit{어리둥절하다} (\textit{ŏritungchŏlhata}) in Korean), which denotes a non-emotional epistemic state.

\subsection{Emotion Concepts and Metaphors}
Moreover, this dataset supports both computational and behavioral research on emotion concepts. Emotion word embeddings could be utilized to conduct computational analyses of lexical gaps, revealing how multilingual language models represent language-specific words. Emotion word embeddings also enable the investigation of foundational questions about emotion structure, such as whether shared dimensions (e.g., valence and arousal) organize the emotion lexicon across languages \citep{niedenthal2008emotion, yik2023relationship}. High-quality translation mappings provided by the dataset, validated by bilingual speakers of both languages, offer a reliable gold standard for machine learning approaches.

Behavioral studies can further examine how conceptual gaps influence the communication or interpretation of those emotions \citep{rissman2023gaps} using the emotion terms in this dataset. Finally, the annotations of multiple meanings in this dataset (\textsc{poly}12--14) supports cross-cultural investigations of emotion-related metaphors. Comparing the identified metaphorical extensions, researchers can assess the universality versus specificity of emotion-related metaphors \citep{kovecses2003metaphor, sauciuc2009role}. 

\section{Conclusion}
The present dataset includes 1,426 Korean words and 1,397 English words, with many-to-many translational mappings documented between the two languages. Using a feature-based annotation approach, the dataset provides a theory-agnostic set of adjectives and verbs that can be selected as emotion words based on criteria drawn from prior studies. The dataset captures both within-language and cross-linguistic features of emotion words. Beyond documenting linguistic variation, the dataset provides several principled bases for selecting emotion words and for comparing emotion-related metaphors across languages. It can serve as a resource for linguistic research, computational analysis of emotion concepts, and empirical studies of emotion. 

\section{Ethical Considerations and Limitations}

The dataset does not involve serious ethical considerations, as it contains no personal or identifiable information. All annotators were assigned anonymized coder IDs.

Despite the dataset's contributions, this study has certain limitations. Although the dataset was constructed to cover more than one language, there are many other languages to investigate for their emotion words. Regarding the annotation, the majority of the data was annotated by a single trained annotator, and the agreement set (about 10\% of adjectives and verbs) has three annotators' judgments adjudicated. This percentage was chosen to prioritize the annotators' effort put into the large number of emotion words in the dataset with complex annotation criteria that required specific training with adequate domain knowledge. Future research could extend the approach to a broader range of languages and include a larger proportion of items annotated by multiple annotators.

\section{Acknowledgements}
We would like to thank the annotators for their dedication and hard work. We also thank anonymous reviewers for their valuable feedback and comments. 

\nocite{*}
\section{Bibliographical References}\label{sec:reference}

\bibliographystyle{lrec2026-natbib}
\bibliography{lrec2026-example}

\bibliographystylelanguageresource{lrec2026-natbib}

\appendix

\section{Modifications to Korean word annotations}
\label{sec:appendix_kor_annot}
For the Korean word annotations, the core instructions were provided in English, consistent with those used for the English word annotations. The sentence frames for \textsc{acpt}1--4 and \textsc{exp}5--8 were translated into Korean as follows (For \textsc{exp}5--8, the sentences and subjects were translated identically to those used in \textsc{acpt}1--4).

For acceptability judgments, the sentence contexts were translated as: ``내가/나는 X고 느낀다.'' (\textsc{acpt}1), ``그들이/은 X고/(이)라고 느낀다.'' (\textsc{acpt}2), ``이것이(은)/저것이(은)/그것이(은) X고/(이)라고 느낀다,' (\textsc{acpt}3), and ``나는 X다.'' (\textsc{acpt}4). As Korean doesn't have a single word equivalent to the English \textit{it}, three different types of subjects were allowed (\textit{이것} `this thing', \textit{저것} `that thing over there', \textit{그것} `that thing') in \textsc{acpt}3. If any one of these subject forms yielded an \texttt{acceptable} judgment, the item was coded as \texttt{acceptable}.

Additionally, annotators were allowed to use whichever word ending felt most natural in each sentence context, as Korean lexical items can appear with different inflectional endings depending on the context. Additional columns were provided for annotators to indicate the most appropriate word form for each frame. For example,\textit{슬프다} (\textit{sŭlpŭta}) feels natural as \textit{슬프다고} in ``내가/나는 X고 느낀다.'' (\textsc{acpt}1) context, and therefore the annotator recorded \textit{슬프다고} in a separate column as a natural word form for \textsc{acpt}1.

\section{Policies on Counting Polysemous Word Entries}
\label{sec:appendix_poly}
Polysemous words are counted as two separate entries under two conditions: (1) has meanings spanning multiple parts of speech (e.g., \textit{worry} as a noun and verb were counted as two separate entries.), or (2) one of the meanings having translations in one language but not in the other (e.g., \textit{needy} was counted as two entries, as one sense had a Korean equivalent and another did not). This approach was adopted to ensure that the total number remained interpretable; otherwise, the sum of the number of words across categories would not correspond to the total number of words in each language. \cref{overlap_pos_korean,overlap_pos_english,translation_overlap_combined} summarizes the number of words overlapping across two conditions.

\begin{table}[ht!]
\centering
\caption{Overlaps Across Parts of Speech in Korean}
\label{overlap_pos_korean}
\begin{tabular}{llccc}
\hline
\textbf{Translation} &  & \textbf{Noun} & \textbf{Verb} & \textbf{Adjective} \\
\hline
\multirow{2}{*}{With} 
& Noun & -- & 0 & 0 \\
& Verb &  & -- & 1 \\
\multirow{2}{*}{Without} 
& Noun & -- & 0 & -- \\
& Verb &  & -- & 0 \\
\hline
\end{tabular}
\end{table}

\begin{table}[ht!]
\centering
\caption{Overlaps Across Parts of Speech in English}
\label{overlap_pos_english}
\begin{tabular}{llccc}
\hline
\textbf{Translation} &  & \textbf{Noun} & \textbf{Verb} & \textbf{Adjective} \\
\hline
\multirow{2}{*}{With} 
& Noun & -- & 35 & 14 \\
& Verb &  & -- & 2 \\
\multirow{2}{*}{Without} 
& Noun & -- & 1 & -- \\
& Verb &  & -- & 0 \\
\hline
\end{tabular}
\end{table}

\begin{table}[ht!]
\centering
\caption{Overlaps Across Translational Status Within Parts of Speech}
\begin{tabular}{lcc}
\hline
\textbf{Part of Speech} & \textbf{Korean} & \textbf{English} \\
\hline
Noun      & 2  & 4  \\
Verb      & 2  & 3  \\
Adjective & 3  & 11 \\
\hline
\end{tabular}
\label{translation_overlap_combined}
\end{table}

\section{Verb Annotation Results}
\label{sec:annot_verb_result}

\subsection{Agreement Scores}

The pairwise kappa scores, along with the aggregated scores across all pairs, of the verb agreement set are presented in \cref{fig_pairwise_kappa_eng_verb,fig_pairwise_kappa_kor_verb}. Overall, annotations of Korean verbs (mean Cohen's $k$ = 0.82; aggregated across all annotation criteria and coders) showed higher reliability than annotations of English verbs (mean Cohen's $k$ = 0.69; aggregated across all annotation criteria and coders), which mirrors the agreement pattern of adjective annotations.

Across individual criteria, questions in the exclusion criteria produced higher agreement (\textsc{excl}9--11; Korean $k$ = 0.88, English $k$ = 0.83). In contrast, annotations on causality criteria exhibited lower and more variable agreement, particularly for Korean verbs across the questions (\textsc{exp}7--8; Korean $k$ = 0.72, English $k$ = 0.48). While the agreement on the ``causing'' question was at an ``almost perfect'' level (mean $k$ = 0.92), the “caused” question (\textsc{exp}7) produced lower agreement (mean $k$ = 0.52) for Korean verbs. This pattern contrasts with that observed for adjectives, where ``causing'' criteria were challenging for both Korean and English annotators (Korean $k$ = 0.46, English $k$ = 0.45). Overall, agreement on exclusion criteria was similarly high for both verbs and adjectives, whereas agreement on causality annotations was higher for verbs than for adjectives.

The agreement patterns of verbs also highlight the subjective nature of emotion concepts, particularly for causality judgments. At the same time, higher agreement on causality annotations for verbs indicates relatively less variation for verbs. The very high level of agreement for ``causing'' criterion suggests that the causative properties of Korean emotion verbs may be more consistently interpretable than those of English emotion verbs. 

\subsection{Annotation Trend}

A summary of annotation trends is in Figure~\ref{fig_annotation_trend_verb}. Annotations in the agreement set were adjudicated using majority vote to produce a single combined dataset for analysis.

In the causality criteria, the general trend of Korean adjectives showing higher acceptance for \textsc{exp}7 but lower acceptance for \textsc{exp}8 than English adjectives remains the same for verbs as well, but the difference between the two criteria was larger (\textsc{exp}7 acceptance: Korean 68.51\%, English 51.15\%; \textsc{exp}8 acceptance: Korean 7.23\%, English 43.08\%). Exclusion criteria (\textsc{excl}9--11) also showed highly consistent rejection in both languages, achieving around 90\% rate of \texttt{no} response. 

Unlike the discrepancy across languages for adjectives, verbs had a similar rate of additional emotion-related meanings being annotated (\textsc{poly}12; 7.76\% in Korean, 7.83\% in English). This rate is similar to that found for Korean adjectives, but remains lower than the rate for English adjectives. These findings may suggest a relatively higher degree of polysemy among English adjectives. 

\begin{figure*}[!ht]
    \centering
    \includegraphics[width=0.87\textwidth]{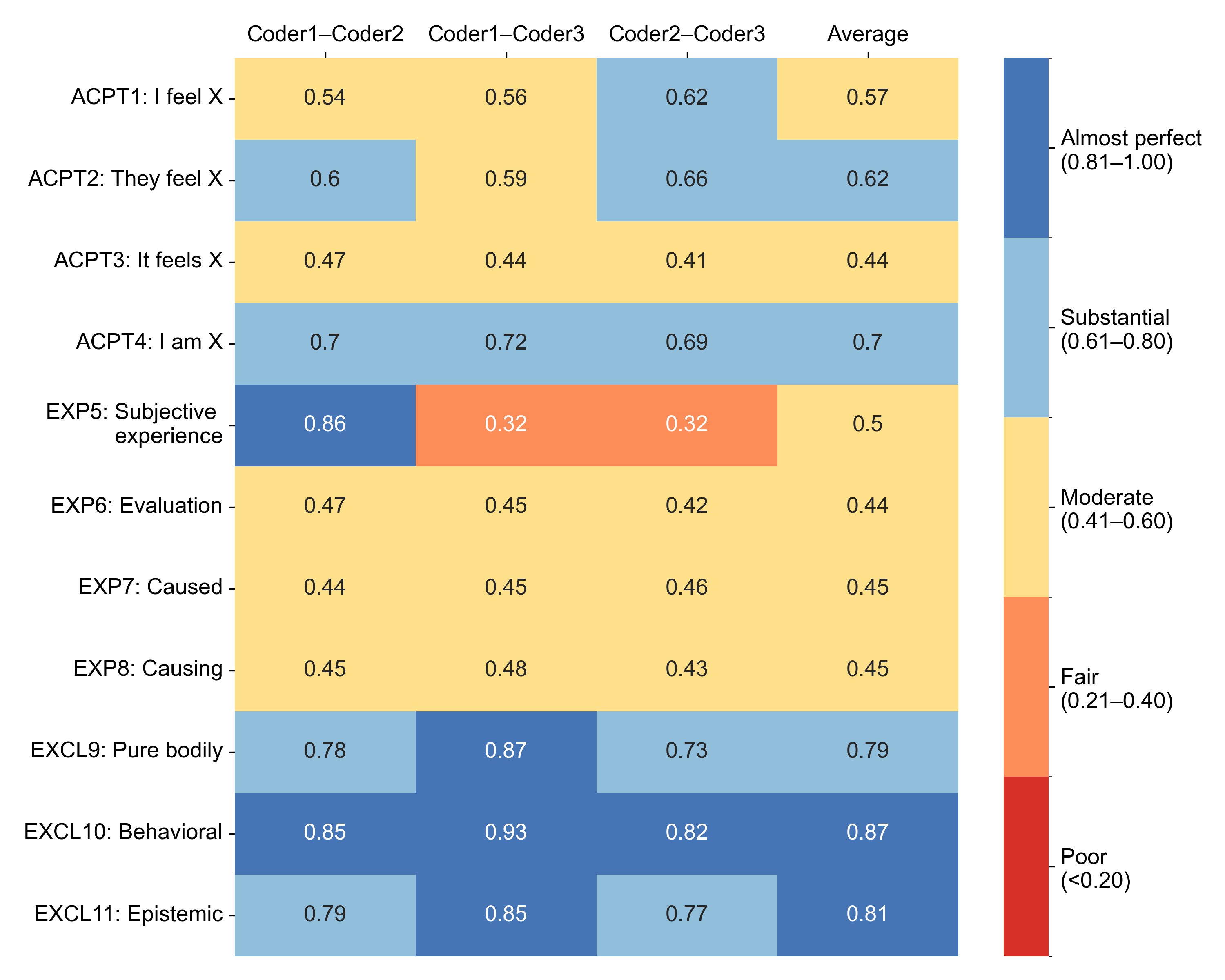}
    \caption{Agreement scores across annotation criteria for the English adjective agreement set.}
    \label{fig_pairwise_kappa_eng_adj}
\end{figure*}

\begin{figure*}[!ht]
    \centering
    \includegraphics[width=0.87\textwidth]{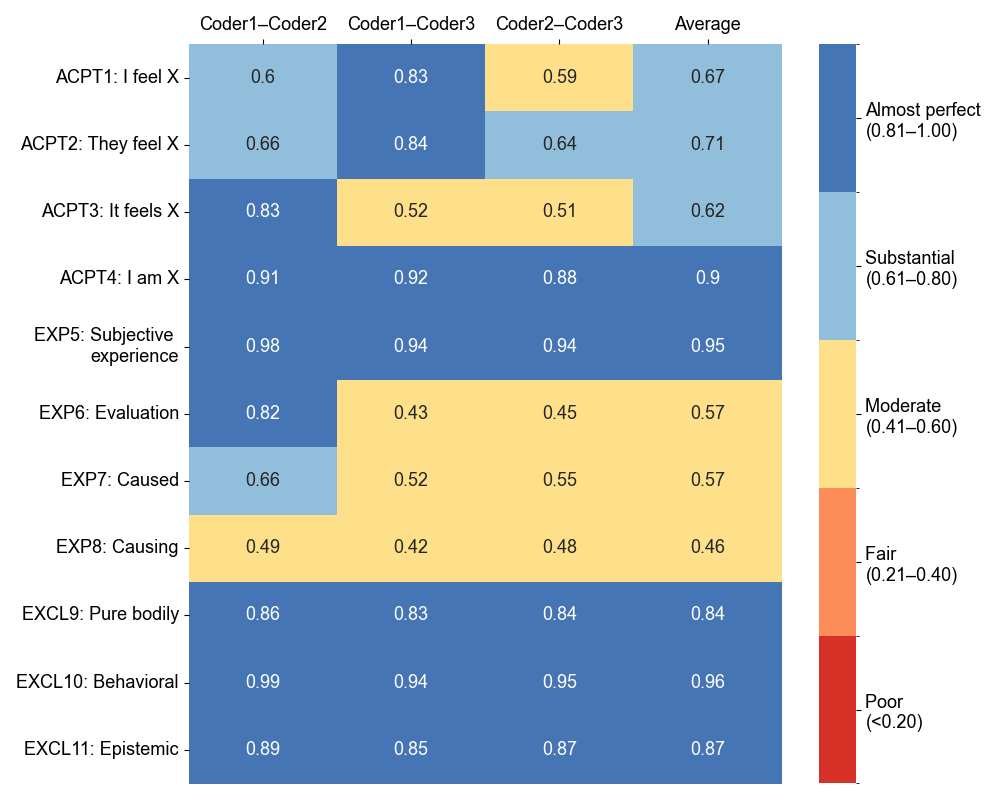}
    \caption{Agreement scores across annotation criteria for the Korean adjective agreement set.}
    \label{fig_pairwise_kappa_kor_adj}
\end{figure*}

\begin{figure*}[!ht]
    \centering
    \includegraphics[width=0.9\textwidth]{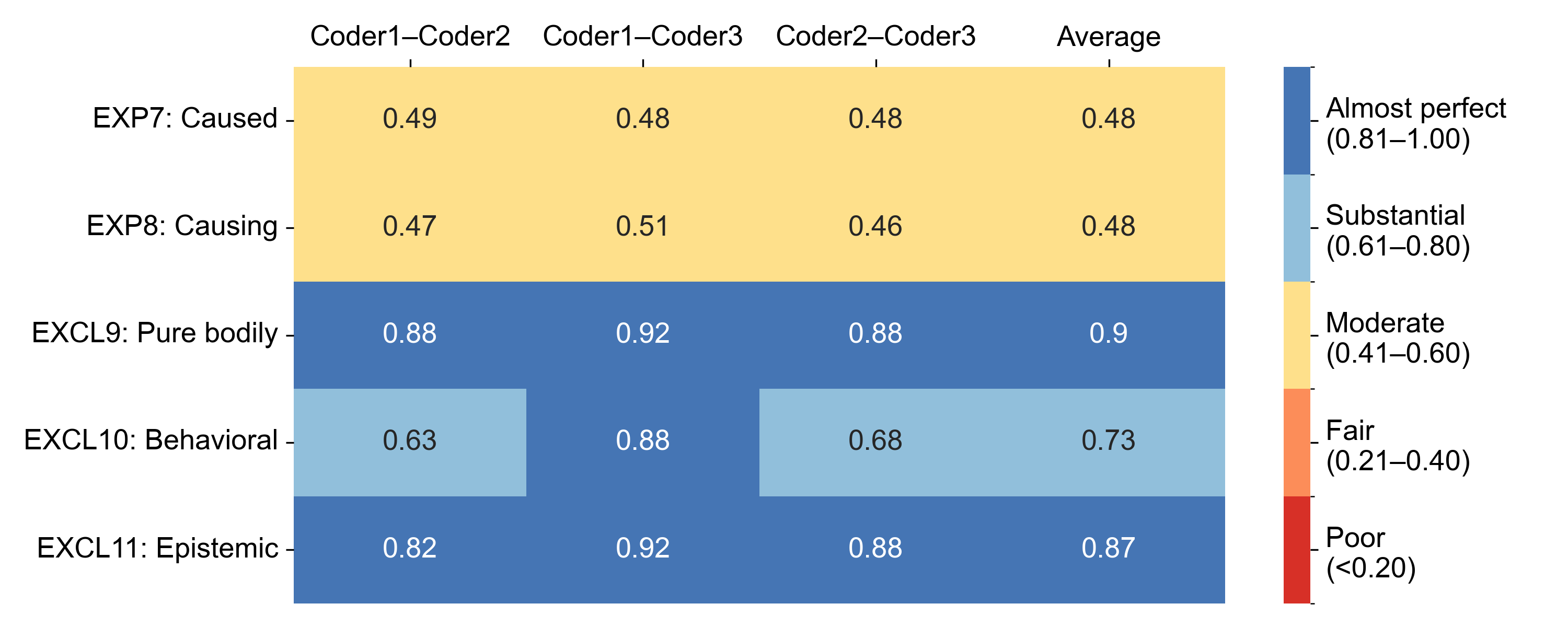}
    \caption{Agreement scores across annotation criteria for the English verb agreement set.}
    \label{fig_pairwise_kappa_eng_verb}
\end{figure*}

\begin{figure*}[!ht]
    \centering
    \includegraphics[width=0.9\textwidth]{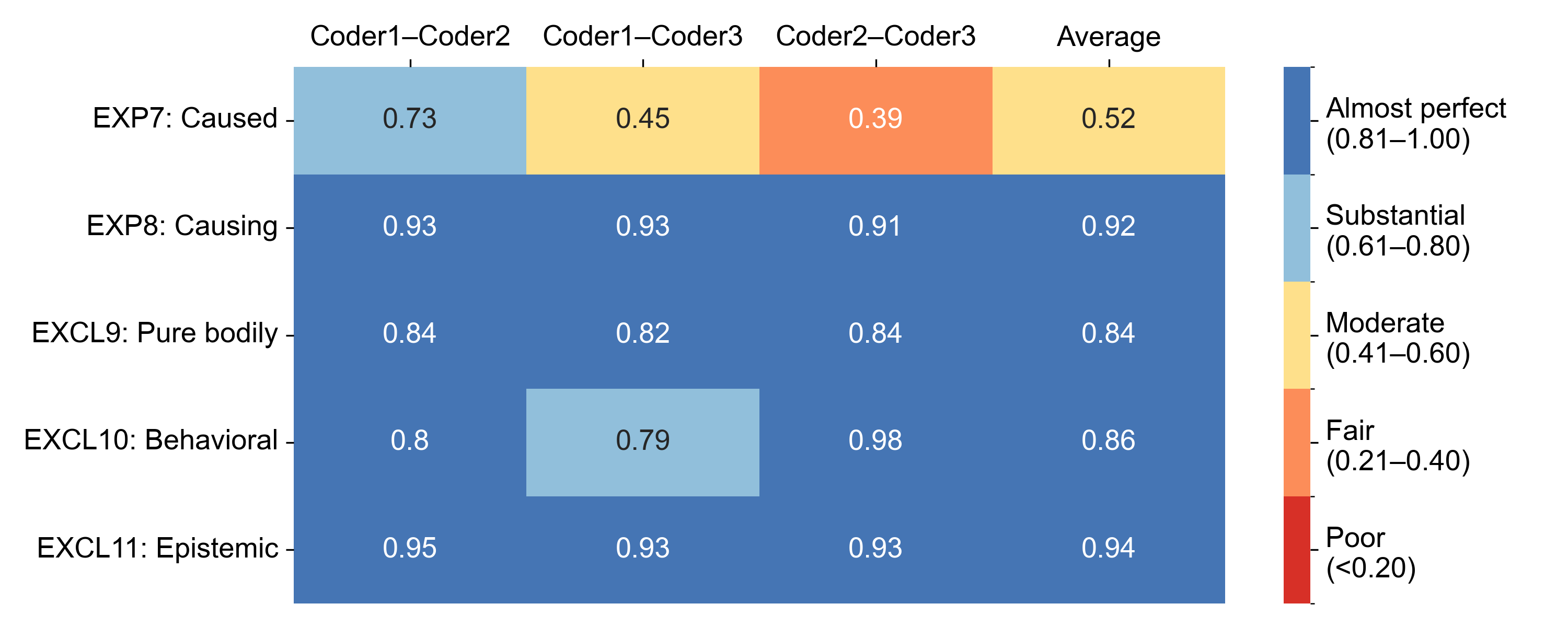}
    \caption{Agreement scores across annotation criteria for the Korean verb agreement set.}
    \label{fig_pairwise_kappa_kor_verb}
\end{figure*}

\begin{figure*}[!ht]
    \centering
    \includegraphics[width=0.95\textwidth]{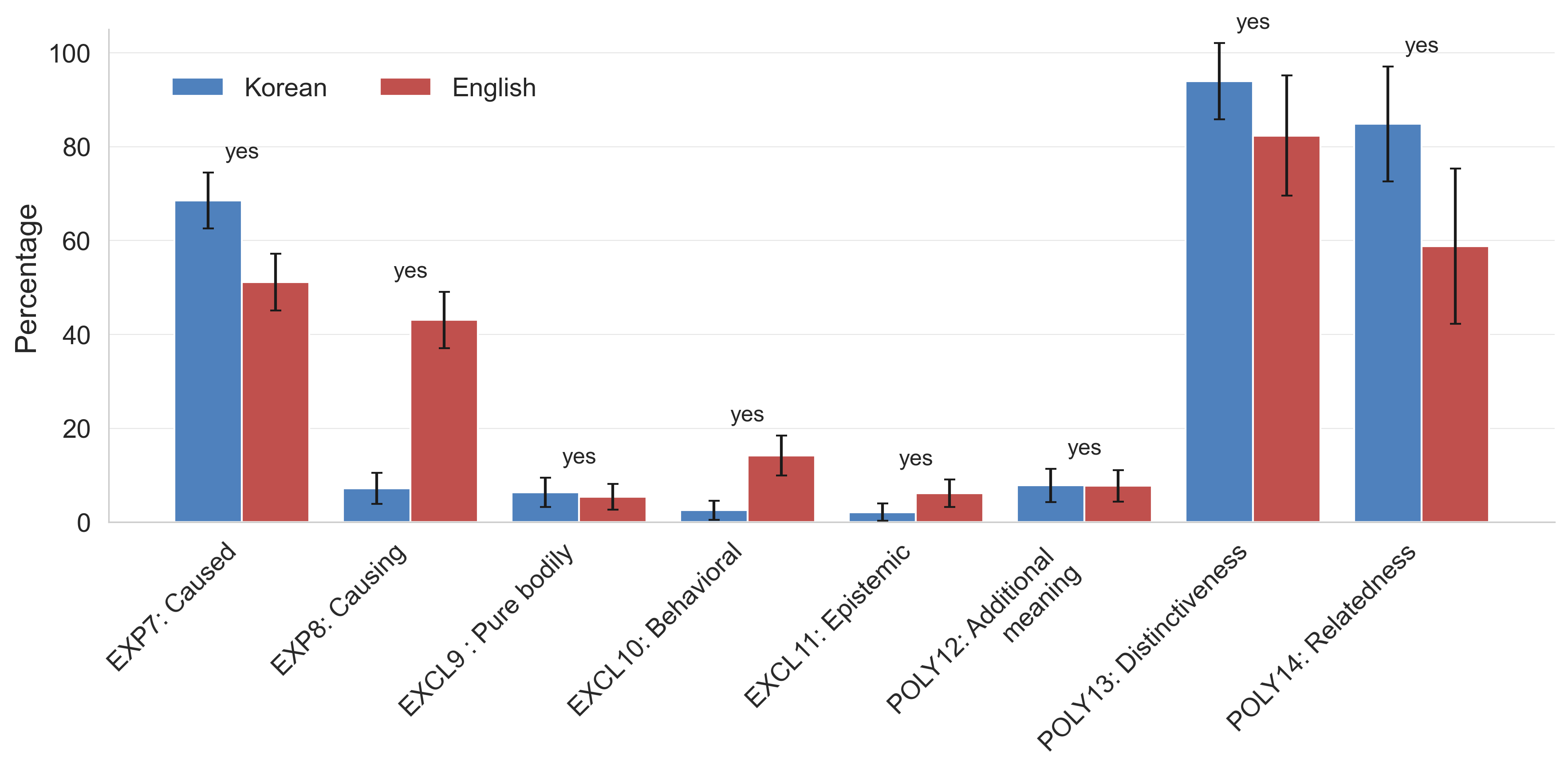}
    \caption{The percentages of annotation values for each criterion for verbs in Korean and English. The percentages for \textsc{poly}12 indicate the ratio of words annotated as having more than one meaning in each language. The percentages for \textsc{poly}13 and \textsc{poly}14 were calculated among items annotated as having more than one meaning in \textsc{poly}12. Error bars indicate 95\% confidence intervals.}
    \label{fig_annotation_trend_verb}
\end{figure*}

\end{document}